\documentclass[letterpaper, 10 pt, journal, twoside]{IEEEtran}
\usepackage{amsmath,amsfonts}
\usepackage{algorithmic}
\usepackage{algorithm}
\usepackage{array}
\usepackage[caption=false,font=footnotesize,labelfont=sf,textfont=sf]{subfig}
\usepackage{textcomp}
\usepackage{stfloats}
\usepackage{url}
\usepackage{verbatim}
\usepackage{graphicx}
\usepackage{cite}
\usepackage{graphicx}
\usepackage{multirow}
\usepackage{amsmath,amssymb,amsfonts}

\begin{document}

\author{Jinghe Yang$^{1}$, Mingming Gong$^{2}$, and Ye Pu$^{1}$%
\thanks{This work was supported by the Melbourne Research Scholarship from the University of Melbourne, the Australian Research Council via grant DE220101527, and the Australian Government via grant AUSMURIB000001.}
\thanks{$^{1}$Jinghe Yang and Ye Pu are with the Department of Electrical and Electronic Engineering, The University of Melbourne, Australia. $^{2}$Mingming Gong is with the School of Mathematics and Statistics, The University of Melbourne, Australia. {\tt\footnotesize jinghey@student.unimelb.edu.au}.}%
}

\title{Knowledge Distillation for Underwater Feature Extraction and Matching via GAN-synthesized Images}
\maketitle

\begin{abstract}
Autonomous Underwater Vehicles (AUVs) play a crucial role in underwater exploration. Vision-based methods offer cost-effective solutions for localization and mapping in the absence of conventional sensors like GPS and LiDAR. However, underwater environments present significant challenges for feature extraction and matching due to image blurring and noise caused by attenuation, scattering, and the interference of \textit{marine snow}. In this paper, we aim to improve the robustness of the feature extraction and matching in the turbid underwater environment using the cross-modal knowledge distillation method that transfers the in-air feature extraction and matching models to underwater settings using synthetic underwater images as the medium. We first propose a novel adaptive GAN-synthesis method to estimate water parameters and underwater noise distribution, to generate environment-specific synthetic underwater images. We then introduce a general knowledge distillation framework compatible with different teacher models. The evaluation of GAN-based synthesis highlights the significance of the new components, i.e. GAN-synthesized noise and forward scattering, in the proposed model. Additionally, VSLAM, as a representative downstream application of feature extraction and matching, is employed on real underwater sequences to validate the effectiveness of the transferred model. Project page: https://github.com/Jinghe-mel/UFEN-GAN.
\end{abstract}


\section{INTRODUCTION}
\IEEEPARstart{R}{ecently}, Autonomous Underwater Vehicles (AUVs) have demonstrated great potential in underwater explorations, replacing human divers in hazardous scenarios. They are capable of performing various tasks, such as underwater facility repairs and inspections. Accurate localization and mapping are imperative for the successful execution of these missions, presenting a significant challenge due to the unavailability of in-air sensors like the Global Positioning System (GPS) and LiDAR. Consequently, AUVs depend on alternative sensors, such as cameras and sonar, for environmental perception and position estimation. Vision-based methods, which utilize cameras as sensing equipment, stand out for their affordability and ease of implementation \cite{sahoo2019advancements}.

Feature point extraction and matching is essential for deriving 3D geometric information from 2D images in applications like Structure-from-Motion (SfM) and Visual Simultaneous Localization and Mapping (VSLAM). Classical methods, such as SIFT \cite{lowe2004distinctive} and ORB \cite{rublee2011orb}, are widely used; for instance, COLMAP \cite{schonberger2016structure} employs SIFT to identify correspondences between image frames for map reconstruction. Similarly, ORB \cite{rublee2011orb} is the front end of the ORB-SLAM \cite{mur2017orb} framework, facilitating camera motion estimation across frames. Recently, learning-based methods have demonstrated significant potential in feature extraction and description \cite{detone2018superpoint, luo2020aslfeat,zhao2023aliked, revaud2019r2d2}, particularly in challenging scenarios such as fast motion \cite{tang2019gcnv2}, underwater turbidity \cite{yang2023knowledge}, and low-light environments \cite{he2023darkfeat}.

Unlike in-air scenarios, where features are distinct, underwater features are often blurred and attenuated due to light absorption and scattering in water. These conditions pose significant challenges for feature extraction and matching in underwater imagery, as the performance of the feature extraction and description deteriorates rapidly with increasing turbidity \cite{oliver2010image}. This degradation adversely affects the robustness of downstream tasks. For instance, in \cite{joshi2019experimental}, the authors compared the in-air feature-based VSLAMs, such as SVO \cite{forster2014svo} and ORB-SLAM2 \cite{campos2021orb}, on underwater image sequences and observed frequent tracking loss under blurriness.

To enhance feature extraction and matching in blurry underwater images, several studies use image restoration techniques to improve image quality and enhance the finding of feature correspondences \cite{cho2017visibility, xie2021variational}. In \cite{hodne2022detecting}, an additional classification model is used to mitigate incorrect feature detections on \textit{marine snow}, which is composed of organic material, detritus, and inorganic particles in water. Another promising approach involves cross-modal knowledge distillation \cite{gupta2016cross}, which transfers feature extraction and matching knowledge learned on clear in-air images to models operating on visually distinct underwater domains. For example, UFEN \cite{yang2023knowledge} employs synthetic underwater images as a medium for this knowledge transfer.

\begin{figure*}[t]
\centering
\includegraphics[width=0.95\linewidth]{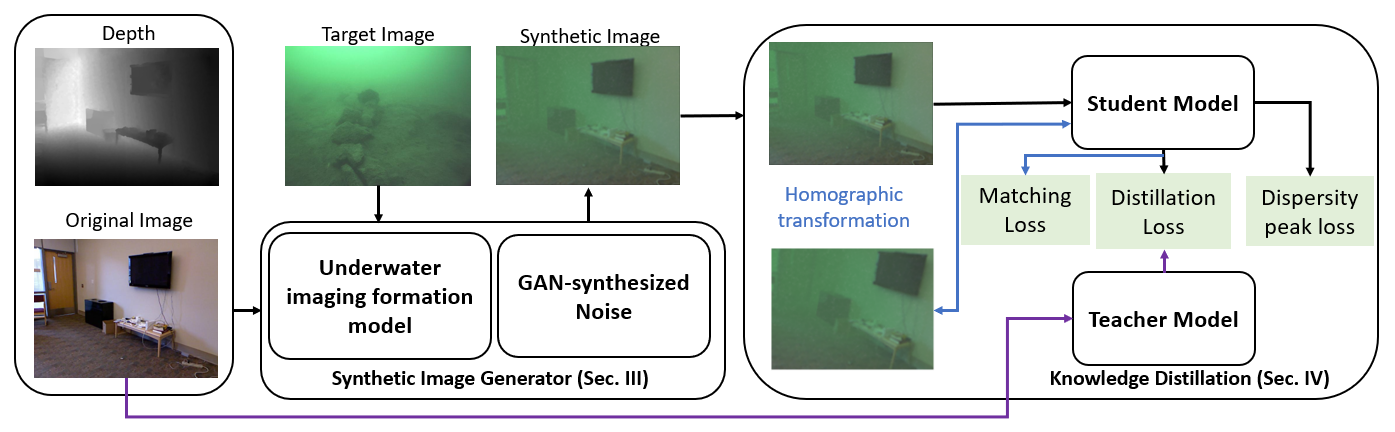}
\caption{\textbf{Overall training framework:} The training framework uses adaptive synthetic images of the target underwater environment as a medium to perform knowledge distillation from the in-air teacher model to the target underwater settings.}
\label{fig:1}
\end{figure*}

The Generative Adversarial Network (GAN) \cite{goodfellow2014generative} has been widely used in underwater image generation tasks to enhance the similarity between synthesized images and target underwater scenes, addressing the challenges of groundtruth collection in underwater environments. This approach serves as a precursor for various downstream tasks, including image restoration \cite{li2017watergan, desai2021ruig, wen2023syreanet, islam2020fast} and depth estimation \cite{hambarde2021uw, zhang2024atlantis}. However, synthetic images for feature extraction should account for not only color appearance in underwater environments but also critical factors such as blurring and image noise, which greatly impact the robustness of feature extraction and matching in underwater settings. While the GAN-based enhancement method proposed in \cite{zheng2023real} improves degraded underwater images and facilitates VSLAM, it may also unintentionally amplify the visibility of floating marine snow, which may lead to a degradation in feature matching performance.

To address these challenges, we propose an adaptive GAN-based synthesis method integrated with a generalized knowledge distillation framework, enabling in-air feature extraction and description models to adapt to underwater scenarios. First, unlike UFEN, which generates the synthetic images using randomly sampled water parameters from various water types \cite{r3}, and injects visually unrealistic Gaussian noise into synthetic images, we propose a GAN-based approach that incorporates adaptive noise distribution and forward scattering into the image formation. We further demonstrate the necessity and effectiveness of these components for improving feature matching transfer performance (see Section~\ref{sec_m1}). Building upon previous work, UFEN, which is specifically designed for transferring SuperPoint \cite{detone2018superpoint}, we propose a general cross-modal knowledge distillation framework that can adapt to various score map-based feature extraction and matching models, supporting different output structures and descriptor formats (see Section \ref{sec_m2}).

The overall framework is illustrated in Fig.~\ref{fig:1}. The key contributions of this paper are as follows: 
\begin{itemize} 
\item We present a GAN-based approach for synthesizing underwater images for specific environments, where the novelty lies in incorporating forward scattering and noise distribution in underwater image synthesis.
\item We propose a general knowledge transfer framework, enabling knowledge distillation from different in-air teacher models to target underwater environments.
\item Our experiments demonstrate that incorporating forward scattering and GAN-synthesized noise significantly improves the transfer performance of feature matching in underwater environments.
\item We validate the effectiveness of our method by integrating the transferred model into the VSLAM framework.
\end{itemize}
\section{Preliminaries}
This section presents the preliminaries of this work, covering the background of the commonly used underwater imaging formation model, learning-based feature extraction and description, and underwater image synthesis.
\subsection{Underwater Imaging Formation Model}
Recent studies have streamlined the underwater imaging model into two components: the direct and the backscattering signal \cite{akkaynak2018revised, chiang2011underwater, peng2015single, yang2024physics}, which can be expressed as:
\begin{equation}
    I_c(x) = \underbrace{J_c(x) \cdot e^{-\beta_c z(x)}}_{\text{Direct Signal}} + \underbrace{B_c^\infty (1 - e^{-\beta_c z(x)})}_{\text{Backscattering}},
    \label{equ:1}
\end{equation}
where \( I_c(x) \) and \( J_c(x) \) denote the intensities at pixel \( x \) in the captured underwater image and the unattenuated image, respectively, across the \( c \in \{R,G,B\} \) color channels. The direct signal describes photons travelling directly from objects to the camera, experiencing energy loss due to scattering and absorption, with the attenuation rate determined by the beam attenuation coefficient \( \beta_c \) and the distance to the camera, \( z(x) \). In contrast, backscattering is caused by the reflection of ambient light $B_c^\infty$ from particles in water. However, the basic model in \eqref{equ:1} is overly simplified, overlooking characteristics of underwater imaging, such as forward scattering and the phenomenon known as marine snow. Marine snow, composed of organic material, detritus, and inorganic particles, plays a crucial role in underwater imaging. Its presence can lead to incorrect feature detections in underwater images \cite{hodne2022detecting} and inconsistent feature descriptions \cite{yang2023knowledge}. Modelling and synthesizing marine snow is particularly challenging due to its highly variable shapes and structures. Some studies address this challenge by employing mathematical models to simulate its behaviour \cite{boffety2012phenomenological, kaneko2023marine}. Furthermore, the motion blur caused by the movement of marine snow adds additional complexity to the synthesis. This paper adopts the model in \eqref{equ:1} as the foundation for synthetic image generation. 
\subsection{Learning-based Feature Extraction and Matching}
In recent years, learning-based feature extraction and matching networks have demonstrated promising performance in identifying correspondences between image frames. Score map-based methods, which generate a score map to identify potential feature points and a descriptor map to assign descriptions to these points, represent one main category. SuperPoint, proposed in \cite{detone2018superpoint}, is trained starting from corner points on synthetic shapes and refined through a homographic adaptation strategy. R2D2 \cite{revaud2019r2d2} jointly learns the repeatability and reliability of features using a discriminative loss, enabling reliable detection and description. ALIKE \cite{zhao2022alike} and ALIKED \cite{zhao2023aliked} utilize a differentiable keypoint detection (DKD) and dispersity peak loss for accurate and repeatable keypoint detection. Meanwhile, learning-based feature extraction methods have been employed to improve feature matching robustness in challenging environments \cite{tang2019gcnv2, yang2023knowledge}, such as UFEN \cite{yang2023knowledge}, which addresses underwater blur by generating synthetic images and applying cross-modal knowledge distillation.
\subsection{Underwater Image Synthesis}
Acquiring underwater groundtruth is challenging for tasks like color restoration and depth estimation. To address this, synthetic underwater images are often used to build large underwater datasets \cite{li2017watergan,hambarde2021uw, desai2021ruig}. The foundational approach leverages the underwater imaging formation model, using in-air RGB-D images with known water-type parameters to generate \cite{yang2023knowledge,hou2020benchmarking}. The GAN method is an approach capable of synthesizing the target underwater environment. WaterGAN \cite{li2017watergan} takes RGB-D images as input to generate synthetic images by combining GAN with the imaging formation model. These synthetic images serve as groundtruth for image restoration tasks. Recent work \cite{zhang2024atlantis} employs a diffusion model to generate diverse underwater scenes using depth images as input, targeting underwater monocular depth estimation tasks.
\begin{figure*}[thp]
\centering
\includegraphics[width=0.9\textwidth]{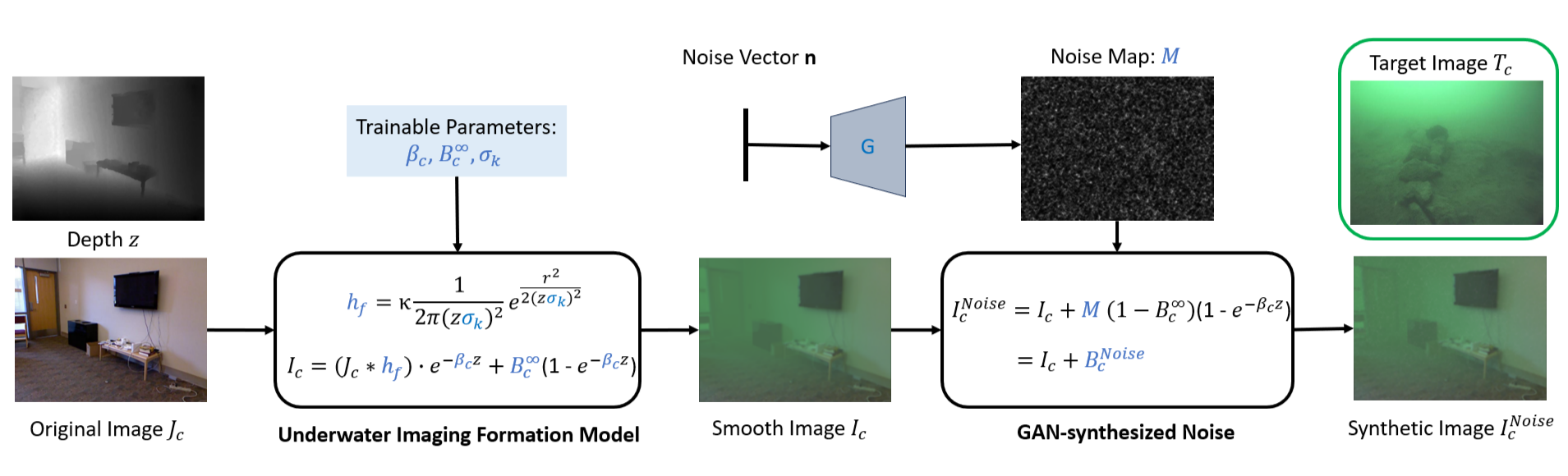}
\caption{\textbf{Synthetic Image Generation Framework:} This framework consists of two components: the underwater imaging formation model considering forward scattering and the GAN-synthesized image noise.}
\label{fig:2}
\vspace{-2mm}
\end{figure*}

\section{GAN-Based Image Synthesis}
\label{sec_m1}
In this section, we propose a model-based GAN approach used for underwater image synthesis, including the refined imaging formation model and the GAN framework.
\subsection{Refined Underwater Imaging Formation Model}
\subsubsection{Forward Scattering}
In the model \eqref{equ:1}, the forward scattering component, which describes light scattering at small angles biased toward the line of sight, is omitted, as its influence on image degradation is less than that of the other two components \cite{schechner2004clear}. However, the existence of forward scattering causes blurriness in images, affecting feature detection and description. This scattering effect can be represented by the Point Spread Function (PSF) applied to the direct signal \cite{jaffe2002computer, schechner2004clear}. Therefore, we consider the following image model with forward scattering in this paper:
\begin{equation}
    I_c(x) = \underbrace{ \left( J_c \ast h_{f} \right)(x)}_{\text{Forward Scattering}} \cdot e^{-\beta_c z(x)} + B_c^\infty \cdot\left( 1 - e^{-\beta_c z(x)} \right),
    \label{equ:2}
\end{equation}
where PSF, denoted as $h_{f}$, is represented as a convolution blur kernel applied to the unattenuated scene image $J_c$. The convolution operator ($*$) is defined as:
\begin{equation} 
\left(J_c \ast h_f \right)(x) = \sum_{x_i \in \mathcal{K}(x)} J_c(x_i) \cdot h_f(x, x_i), 
\label{equ:conv} 
\end{equation}
with $\mathcal{K}(x)$ representing the set of pixels in the kernel centered at $x$. The PSF can be expressed as a Gaussian function \cite{zhu2010image}:
\begin{equation}
     h_{f}(x, x_i) = \kappa(x) \cdot \exp \left(-\frac{r(x, x_i)^2}{2\sigma(z(x))^2}\right),
    \label{equ:3}
\end{equation}
where $r(x, x_i)$ represents the pixel distance from $x_i$ to the pixel $x$ within the Gaussian kernel, and $\sigma(z(x))$ is the standard deviation dependent on the depth $z(x)$. The normalization factor $\kappa(x)$ ensures that:
\begin{equation}
     \sum_{x_i \in  \mathcal{K}(x)}h_f(x, x_i) = 1.
     \label{equ:4}
\end{equation}

Given that the radius of the Point Spread Function (PSF) in water increases linearly with distance, we define \(\sigma(z(x)) = \sigma_k \cdot z(x)\), where \(\sigma_k\) is a trainable scalar factor controlling the effect of the PSF. We use a constant kernel size of $11$ for overall images and perform normalization based on \eqref{equ:4}.

\subsubsection{Image Noise}
Image noise from marine snow affects feature extraction. A recent study \cite{hodne2022detecting} demonstrated that marine snow significantly impairs feature extraction and matching performance. Moreover, the irregular motion of marine snow leads to inconsistent feature descriptions, thereby reducing feature-matching accuracy. Our synthesis process thus accounts for the image noise caused by marine snow, which consists of suspended particles that are part of backscattering and accumulate with scene depth. We model this noise as an additive perturbation to the pure backscattering background, similar to \cite{hodne2022detecting}. We define the noise map \( M \), which has the same dimensions as the image, applied to the pure backscattering background \( B_c^\infty \). The backscattering background can be viewed as being partially filled with marine snow, represented as $B_c^\infty \bigl(1 - M(x)\bigr) + M(x)$. Here, \( M(x) \in [0, 1] \) represents a weight value at each pixel location \( x \). We denote the additional image noise as \( B_c^{Noise} \), formulated by:
\begin{equation}
 B_c^{Noise}(x) = M(x)(1-B_c^\infty)(1 - e^{-\beta_c z(x)})
 \label{bcnoise}
\end{equation}

\subsubsection{Final Imaging Model}
Considering both forward scattering and image noise in the underwater environment, $I^{Noise}_c$ represents the final obtained underwater image. We propose to use the following imaging formation model:

\begin{equation}
    I^{Noise}_c(x) = I_c(x) +  B_c^{Noise}(x),
    \label{final model}
\end{equation}
where \( I_c \) and \( B_c^{Noise} \) are defined in \eqref{equ:2} and \eqref{bcnoise}, respectively.

\subsection{GAN Training Framework}
\subsubsection{Framework Overview}
Natural underwater environments vary significantly in turbidity levels, color appearance and the distribution of the suspended particles. We propose a model-based GAN method to simulate the target environment to enhance knowledge distillation results for a specific environment. Underwater image synthesis for feature extraction purposes differs from restoration tasks \cite{li2017watergan, wen2023syreanet, liu2022adaptive}, which focuses on color recovery while neglecting the effects of forward scattering and marine snow that influence feature extraction and matching. Moreover, it is crucial to avoid changing the positions of feature points during synthesis. Therefore we combine the proposed physical model with a neural-based generator in our generation process. The GAN generation framework is shown in Fig.~\ref{fig:2}.
\subsubsection{GAN Image Generation}
The GAN image generation consists of two components: the first generates images purely based on the physical imaging formation model, where the parameters \( \sigma_k \), \( \beta_c \), and \( B_c^\infty \) in the imaging formation model \eqref{equ:2} are trainable. This component produces smooth synthetic images without noise (denoted as \( I_c \)). The second component generates an estimate of the noise present in underwater images, including electronic noise from devices, uneven particle distributions in water, and marine snow. Since these types of noise are challenging to represent using a parameterized model, we use a neural-based generator \( \mathbf{G}: \mathbb{R}^{N} \to \mathbb{M} \) to generate the noise map \( M \sim \mathbb{M} \) in \eqref{bcnoise}, where \(\mathbf{n} \in \mathbb{R}^{N} \) is the random input vector of dimension $N$. This noise is then added to the smooth synthetic images to produce the final synthetic image (\(I^{Noise}_c\)). To encourage the generated synthetic images to closely resemble the target underwater images, we adopt a modified LSGAN \cite{mao2017least}. The discriminator \( \mathbf{D}: \mathbb{I} \to [0, 1] \) takes both real images \( T_c \sim \mathbb{I} \) and synthetic images \( I^{Noise}_c \sim \mathbb{I} \) as input, where $\mathbb{I}$ represents the image space. The training objectives are defined as follows:
\begin{equation}
\begin{aligned}
\mathcal{L}_{gen} &= \mathbb{E}_{I^{Noise}} \left[ (\mathbf{D}(I^{Noise}_c) - 1)^2 \right], \\
\mathcal{L}_{disc} &= \mathbb{E}_{I^{Noise}} \left[ \mathbf{D}(I^{Noise}_c)^2 \right] +  \mathbb{E}_{T_c } \left[ (\mathbf{D}(T_c) - 1)^2 \right],
\end{aligned}
\label{gan_loss}
\end{equation}
where \(\mathcal{L}_{gen}\) is employed to update the trainable parameters in the physical imaging formation model ($\sigma_k, \beta_c, B_c^\infty$) as well as the image noise generator $\mathbf{G}$, while \(\mathcal{L}_{disc}\) denotes the loss function for the discriminator \(\mathbf{D}\).
\begin{figure}[t p]
\centering
\includegraphics[width=0.95\linewidth]{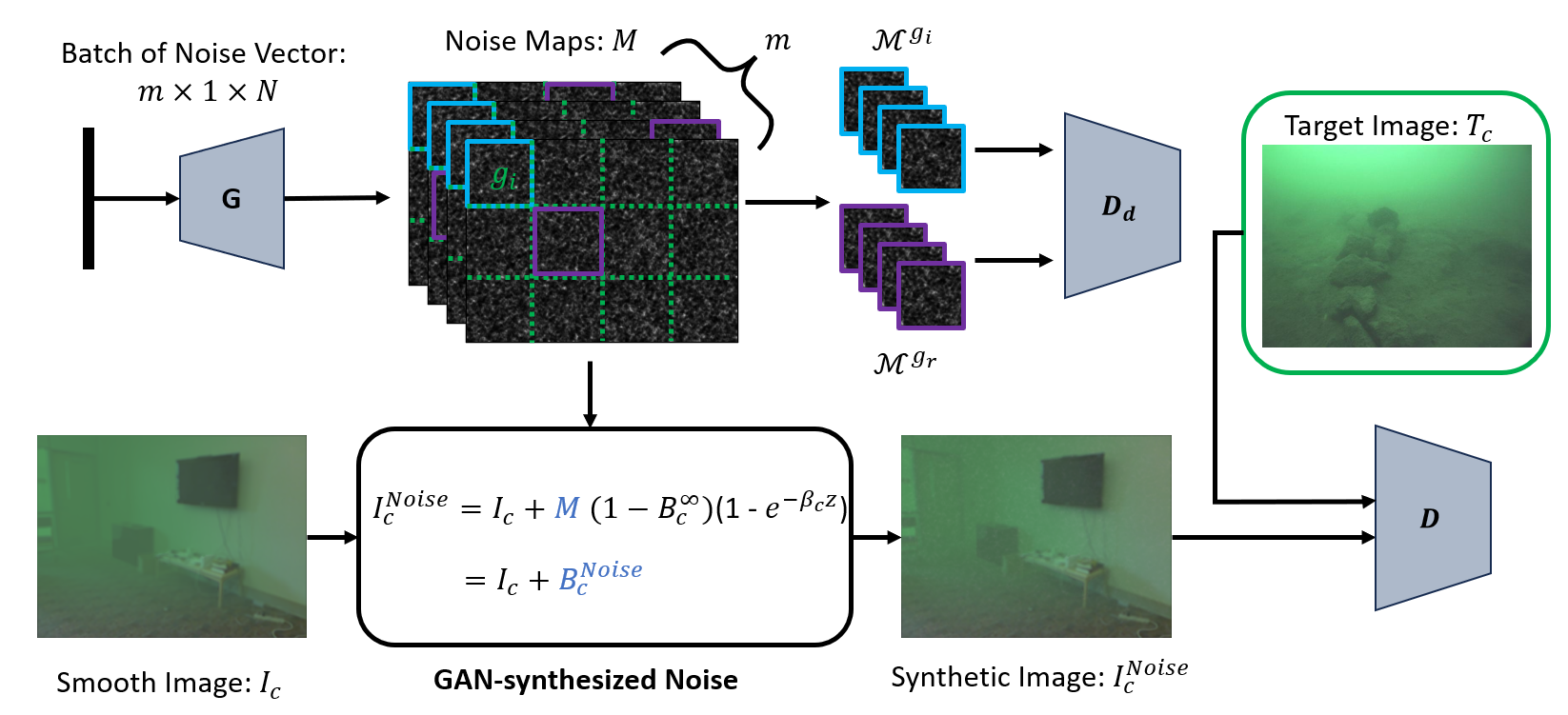}
\caption{\textbf{Training the Noise Generator \( \mathbf{G} \):} The discriminator \( \mathbf{D_d} \) is to ensure a uniform distribution of the noise map.}
\vspace{-4mm}
\label{fig:3}
\end{figure}
\subsubsection{Evenly Distributed Noise Constraint}
Meanwhile, to ensure that the generated noise map \( M \) from $\mathbf{G}$ is evenly distributed across images, we introduce an additional discriminator \( \mathbf{D_d}: \mathbb{R}^{m \times w \times h} \to [0, 1] \), in which \( m \) is the batch size, $w$ and $h$ are the grid size in width and height. Let \( \mathcal{G} \) denote the set of grid locations, uniformly sampled from the noise map $M$. For each batch of noise samples \( \mathcal{M} = \{M_1, M_2, \dots, M_m \} \), with each \( M_j\sim\mathbb{M} \) $(j = 1, \dots, m)$, the discriminator compares patches sampled from the same grid location across the batch with patches sampled from random grid locations across all batch samples. Let \( g_i \in \mathcal{G} \) be the grid location on patch \( i \) ($i = 1, \dots, \frac{HW}{hw}$), $W$ and $H$ are the image size in width and height. The batch of patches from the same grid location is denoted as:
\begin{equation}
\mathcal{M}^{g_i} = \{ M_1^{g_i}, M_2^{g_i}, \dots, M_m^{g_i} \}, \quad g_i \in \mathcal{G},
\end{equation}
where \( M_j^{g_i} \) is the patch sampled from grid location \( g_i \) on the noise map \( M_j \).
Similarly, the batch of patches sampled from random grid locations across all samples is denoted as:
\begin{equation}
\mathcal{M}^{g_r} = \{ M_1^{g_1}, M_2^{g_2}, \dots, M_m^{g_m} \}, \ \{g_1, \dots, g_m \} \in \mathcal{G}.
\end{equation}

This approach ensures that noise is evenly distributed across both individual images and batches. We denote $\mathcal{M}^{g_r}$  as the batch of patches sampled from random grid locations and $\mathcal{M}^{g_i}$ as the batch of patches sampled from fixed grid locations across samples. The distribution loss is defined as:
\begin{align}
\mathcal{L}^{dist}_{gen} &= \mathbb{E}_{\mathcal{M}^{g_r} } \left[ (\mathbf{D_d}(\mathcal{M}^{g_r}) - 1)^2 \right], \\
\mathcal{L}^{dist}_{disc} &= \mathbb{E}_{\mathcal{M}^{g_r}} \left[ \mathbf{D_d}(\mathcal{M}^{g_r})^2 \right] +  \mathbb{E}_{\mathcal{M}^{g_i} } \left[ (\mathbf{D_d}(\mathcal{M}^{g_i}) - 1)^2 \right] \notag,
\end{align}
where $\mathcal{L}^{dist}_{disc}$ is the loss function for the discriminator $\mathbf{D_d}$. The final learning objective of the noise generator $\mathbf{G}$, combined with $L_{gen}$ in \eqref{gan_loss}, is to minimize:
\begin{align}
\mathcal{L}_{g} &= \mathcal{L}_{gen} + \lambda \mathcal{L}^{dist}_{gen},
\end{align}
where \( \lambda \) is a weight parameter to balance the losses. The flow diagram for training $\mathbf{G}$ is shown in Fig.~\ref{fig:3}.
\section{Adaptive knowledge distillation framework for underwater feature extraction}
\label{sec_m2}
After generating synthetic underwater images, we apply cross-modal knowledge distillation, where the supervision for feature matching is provided by the teacher model on in-air images to train the student model on the corresponding synthetic underwater images. Specifically, the teacher model takes in-air images as input, whereas the student model takes the corresponding synthetic images as input.
\subsection{Distillation Loss}
We perform the knowledge distillation step after generating synthetic images for a specific environment. The training framework for this process resembles that described in \cite{yang2023knowledge}, which includes both the distillation loss from the extracted feature score map and the matching loss. We denote the feature score map and the corresponding descriptor map generated by the teacher model as $\mathcal{X}^{Tea}$ and $\mathcal{D}^{Tea}$, and those by the student model as $\mathcal{X}^{Stu}$ and $\mathcal{D}^{Stu}$. The knowledge distillation loss for the feature score map is defined as:
\begin{equation}
    \mathcal{L}_{KD} = f_{\mathcal{X}}(\mathcal{X}^{Stu}, \mathcal{X}^{Tea}) + \alpha_{KD} f_{\mathcal{D}}(\mathcal{D}^{Stu}, \mathcal{D}^{Tea}),
\end{equation}
where $f_{\mathcal{X}}(\cdot)$ and $f_{\mathcal{D}}(\cdot)$ represent the similarity loss functions used to compare the similarities between $\mathcal{X}^{Tea}$ and $\mathcal{X}^{Stu}$, and $\mathcal{D}^{Tea}$ and $\mathcal{D}^{Stu}$, respectively (e.g., $\mathcal{L}_2$ loss and Kullback-Leibler loss). $\alpha_{KD}$ is a weight factor.

\subsection{Dispersity Peak Loss}
To improve score precision at detected feature points, we apply the dispersity peak loss \cite{zhao2022alike}. Given an image \(I_c^{Noise}\), \(N^{Stu}\) feature points \(x_i^{Stu}\) (\(i \in \{1, \dots, N^{Stu}\}\)) are extracted using \(\mathcal{X}^{Stu}\). For each feature point \( \mathbf{x}^{Stu}_i \), a local patch of size \( S \times S \) centered at \( \mathbf{x}^{Stu}_i\) is defined and the \textit{Softmax} is applied to the score map within these patches. The pixel locations in the patch centered at $x_i^{Stu}$ are denoted as $x_{i,j}^{Stu}$, where $j \in \{1, 2, \dots, S^2\}$, and the scores after \textit{Softmax} at these pixels are denoted as $s(x_i^{Stu})$. $r(x_i^{Stu}, x_{i,j}^{Stu})$ is the distance function, same to that in \eqref{equ:3}. Dispersity peak loss $\mathcal{L}_{peak}$ is defined as:
\begin{equation}
    \mathcal{L}_{peak} =\frac{1}{N^{Stu}} \sum^{N^{Stu}}_{i=1} \sum^{S^2}_{j=1 } r(x_i^{Stu}, x_{i,j}^{Stu}) s(x_{i,j}^{Stu}).
\end{equation}

\subsection{Matching Loss}
To ensure the matching performance of the descriptors, we use a matching loss $\mathcal{L}_{M}$ to minimize the descriptor distance between matching feature points and maximize the descriptor distance between non-matching feature points. We apply a homographic transformation $\mathcal{H}$ to the synthetic images $I^{Noise}_c$ to obtain the image pair denoted as $I_c^\mathcal{H}$. $N^\mathcal{H}$ matching feature are then extracted from $I_c^\mathcal{H}$, denoted as $x_j^\mathcal{H}$, where $j \in \{1, 2, \dots, N^\mathcal{H}\}$. The corresponding descriptors of the points, $x^{stu}_{i}$ and $x^\mathcal{H}_{j}$ are denoted as $d_i^{Stu}$ and $d_j^\mathcal{H}$. The matching loss is defined as:
\vspace{-2mm}
\begin{equation}
 \mathcal{L}_M = \frac{1}{Z^2 N^{\mathcal{H}}} \sum_{i=1}^{N^{Stu}} (p_{i}^2 + n_{i}^2),
\end{equation}
where, for $j \in \{1, 2, \dots, N^\mathcal{H}\}$,
\begin{equation}
\begin{aligned}
& p_i = \max(0, \text{dist}(d_i^{Stu}, d_j^\mathcal{H}) - P), \\
& n_i = \max(0, Q - \text{dist}_n(d_i^{Stu}, d_j^\mathcal{H})),
\end{aligned}
\label{equ_match}
\end{equation}
where \(Z\) is a scaling factor, equal to 1 for the float descriptor or to the descriptor dimension in the binary case. In \eqref{equ_match}, \(\text{dist}(\cdot)\) represents the descriptor distance between the matching descriptors \(d_i^{Stu}\) and the corresponding descriptor in the set \(d_j^\mathcal{H}\), while \(\text{dist}_n(\cdot)\) represents the minimal descriptor distance (e.g., \(L_2\) distance) to the non-matching descriptors in \(d_j^\mathcal{H}\). $P$ and $Q$ are two distance margins for the matching and non-matching descriptors, respectively. To accommodate the varying descriptor dimensions and formats across teacher models, we calculate the binary descriptor distance using:
\begin{equation}
\text{dist}(d_i^{Stu}, d_j^\mathcal{H}) = \frac{1}{2} (Z - d_i^{Stu} \cdot d_{j}^{\mathcal{H}}).
\end{equation}

Following UFEN \cite{yang2023knowledge}, we use the straight-through estimator to avoid zero gradients during backpropagation in binarization.

\subsection{Final Loss}
The weights $\gamma_1$ and $\gamma_2$ are used to balance the loss terms.
\begin{equation}
    \mathcal{L} = \mathcal{L}_{KD} + \gamma_1\mathcal{L}_{peak} + \gamma_2\mathcal{L}_M
\end{equation}
\section{Experiments}
In this section, we first evaluate the performance of the GAN-synthesized images through quantitative and qualitative analyses. Next, we assess models distilled from different teacher models, including SuperPoint and ALIKE, in terms of feature extraction and matching. Finally, we evaluate the transferred model on the downstream VSLAM task.
\subsection{Implementation Details}
The training dataset for the GAN synthesis process consists of 1449 images from the NYU-v2 dataset \cite{silberman2012indoor}, with 1149 images used for training and 300 for testing. The depth values are normalized between 0.5m and 5m to ensure scene consistency. Inspired by the adversarial disentanglement training method in \cite{pizzati2023physics}, we first perform the parameter estimation of the formation model in \eqref{equ:2} by freezing the noise generator and optimizing only the physical parameters in the formation model. The training is conducted with a batch size of 4 for 10k iterations. The learning rate is set to \(10^{-3}\) for the trainable parameters and \(10^{-4}\) for the discriminator \(\mathbf{D}\), with updates occurring every 5 iterations. Then, the noise distribution estimation is performed with a batch size of 4 for 20k iterations. The learning rate is set to \(10^{-5}\) for both the generator \(\mathbf{G}\) and the reinitialized discriminators \(\mathbf{D}\) and \(\mathbf{D_d}\), with updates every 10 iterations, and \(\lambda\) is set to 0.1. The structure of the discriminators follows the multi-scale discriminators proposed in \cite{wang2018high}. The generator, which is based on \cite{huang2018multimodal}, replaces the encoder with a fully connected layer to adjust input random noise with size \(N=10\) to the input size of the residual blocks and modifies the output layer to employ a \textit{Sigmoid} activation function. In the knowledge distillation step, we continue using the dataset from the previous stage, training for 30 epochs. The learning rate is set to \(10^{-5}\) for ALIKE and \(10^{-6}\) for SuperPoint. \(\gamma_1\), \(\gamma_2\), and \(\alpha_{KD}\) are set to 1, 1, and 0.01, respectively.

\begin{table}[t]
\centering
\caption{Quantitative Evaluation on the Lake-C and EASI-T Sequences}
\begin{tabular}{|c|c|c|c|c|c|c|}
\hline
\multirow{2}{*}{\textbf{Models}} & \multicolumn{2}{c|}{\textbf{Lake-C}} & \multicolumn{2}{c|}{\textbf{EASI-T}} & \multicolumn{2}{c|}{\textbf{AQUALOC}} \\ \cline{2-7} 
 & \textbf{M. N} & \textbf{M. R} & \textbf{M. N} & \textbf{M. R} & \textbf{M. N} & \textbf{M. R}  \\ \hline
 
ALIKE & 258.0 & 0.877 & 139.5 & 0.641 & 212.2 & 0.756 \\ \hline
\textbf{ALIKE (KD)} & \textbf{264.0} & \textbf{0.929} & \textbf{177.6} & \textbf{0.769} & \textbf{214.0} & \textbf{0.817} \\ \hline
SuperPoint & 153.3 & 0.592 & 130.1 & 0.550 & 306.5 & 0.904\\ \hline
\textbf{SP (KD)} & \textbf{275.2} & \textbf{0.904} & \textbf{191.3}&  \textbf{0.780} & \textbf{335.4} & \textbf{0.952} \\ \hline
UFEN & 244.2 & 0.807 & 146.0 & 0.576 & 300.1 & 0.890 \\ \hline
\textbf{SP (KD-B)} & \textbf{246.6} & \textbf{0.861} & \textbf{172.9} & \textbf{0.744} & \textbf{333.2} & \textbf{0.943} \\ \hline
\end{tabular}
\label{tabel1}
\vspace{-4mm}
\end{table}

\begin{figure}[b]
\centering
\vspace{-4mm}
\includegraphics[width=0.115\textwidth]{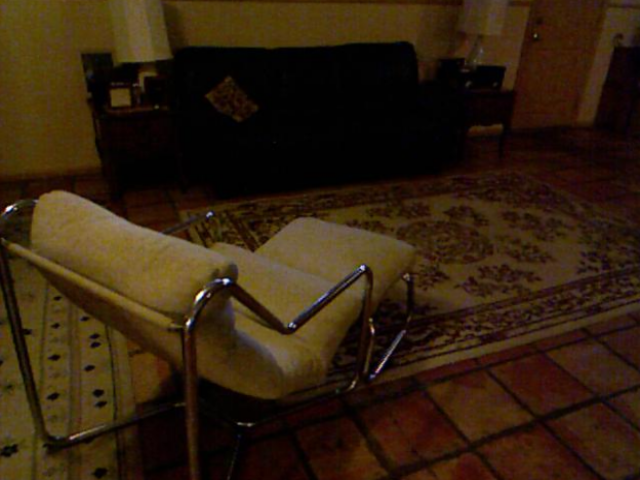}%
\includegraphics[width=0.115\textwidth]{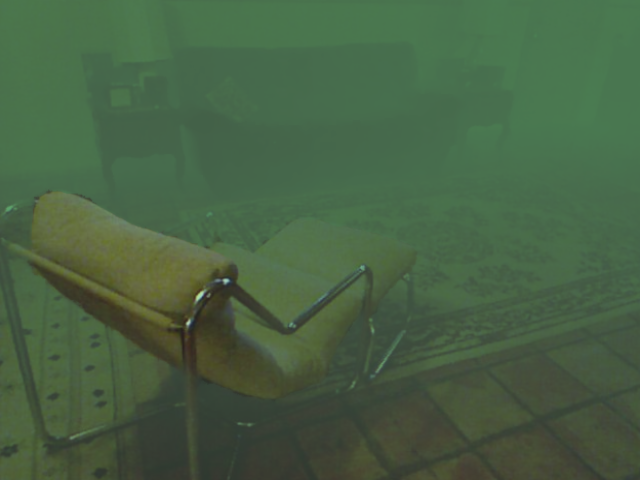}%
\includegraphics[width=0.115\textwidth]{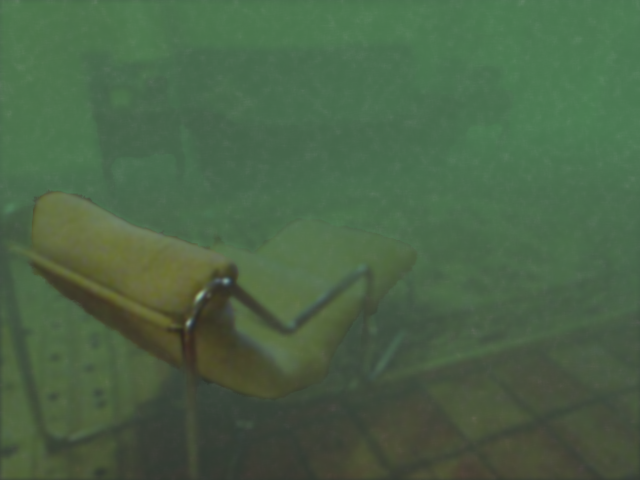}%
\includegraphics[width=0.115\textwidth]{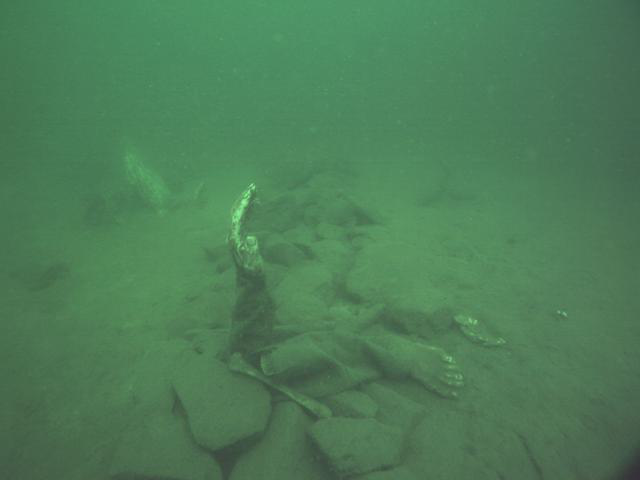}%

\includegraphics[width=0.115\textwidth]{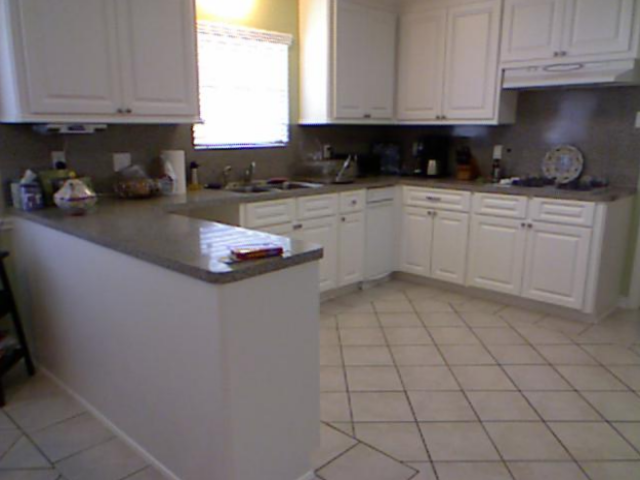}%
\includegraphics[width=0.115\textwidth]{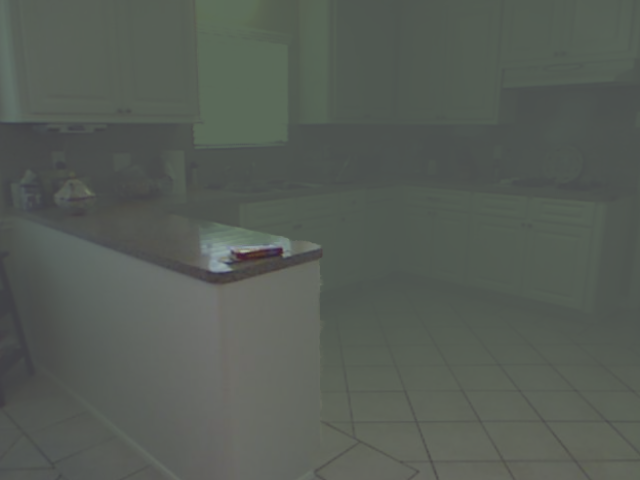}%
\includegraphics[width=0.115\textwidth]{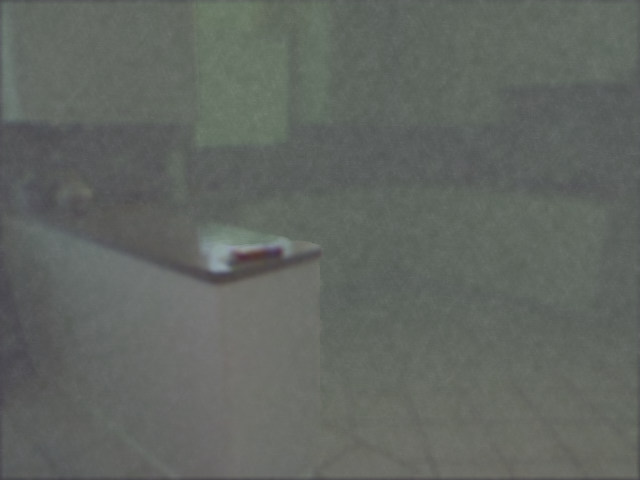}%
\includegraphics[width=0.115\textwidth]{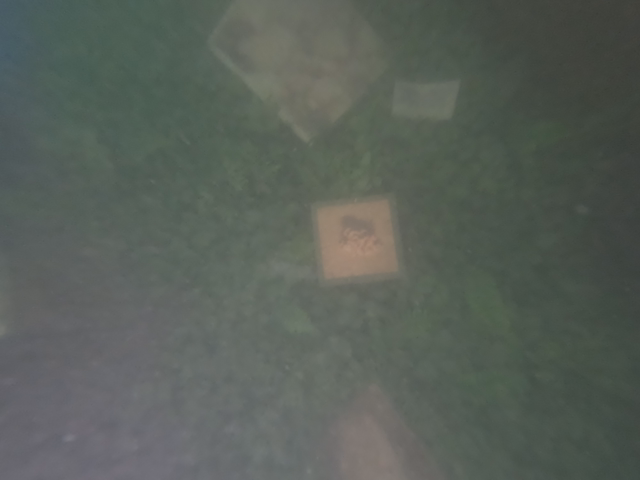}%

\begin{minipage}[t]{0.115\textwidth}
\centering
\includegraphics[width=\linewidth]{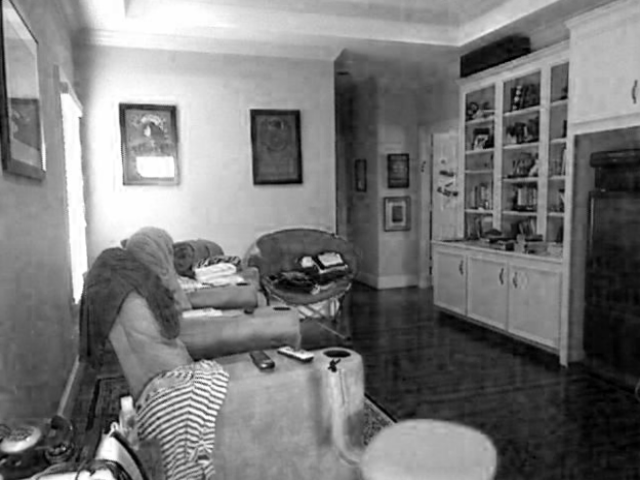}\\
\footnotesize $J_c$
\end{minipage}%
\begin{minipage}[t]{0.115\textwidth}
\centering
\includegraphics[width=\linewidth]{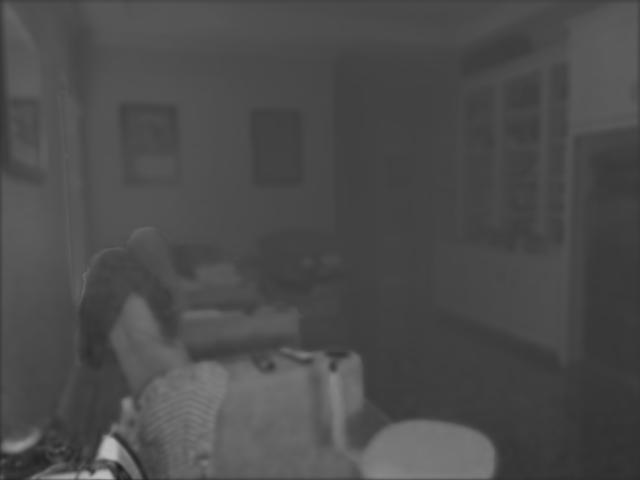}\\
\footnotesize $I_c$
\end{minipage}%
\begin{minipage}[t]{0.115\textwidth}
\centering
\includegraphics[width=\linewidth]{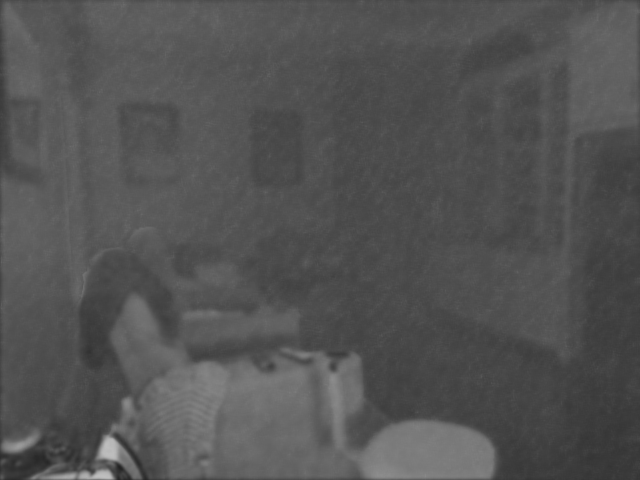}\\
\footnotesize $I_c^{\text{Noise}}$
\end{minipage}%
\begin{minipage}[t]{0.115\textwidth}
\centering
\includegraphics[width=\linewidth]{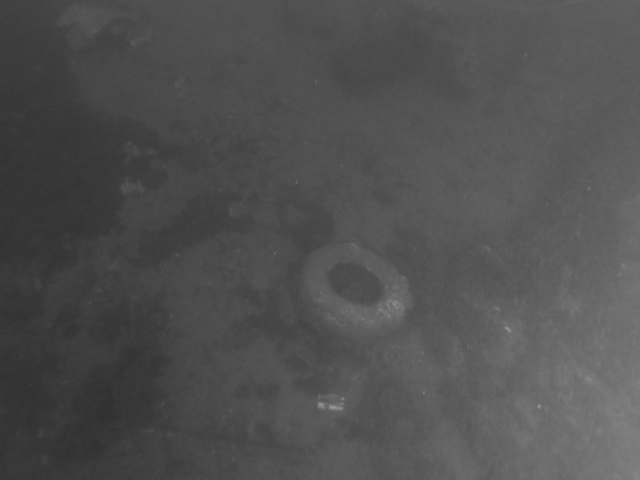}\\
\footnotesize $T_c$ (real)
\end{minipage}
\caption{\textbf{Qualitative Results on Synthetic Images:} Synthetic images are generated based on real target underwater images.}
\label{fig:4}
\end{figure}

\subsection{GAN Synthetic Results}
\subsubsection{Target Underwater Image Synthesis}
\begin{figure}[b]
\vspace{-4mm}
\centering
\subfloat[FID scores]{%
    \includegraphics[width=0.23\textwidth]{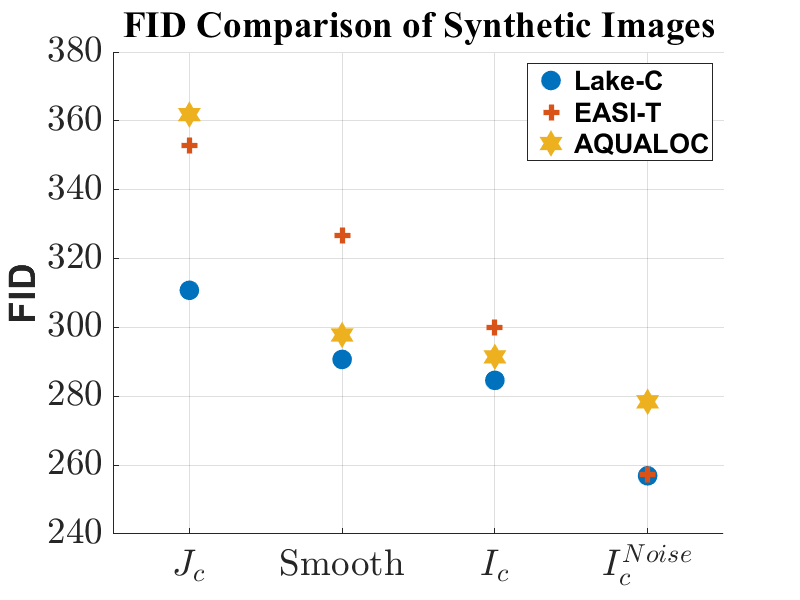}
    \label{fig:5_a}
}%
\subfloat[t-SNE visualization]{%
    \includegraphics[width=0.23\textwidth]{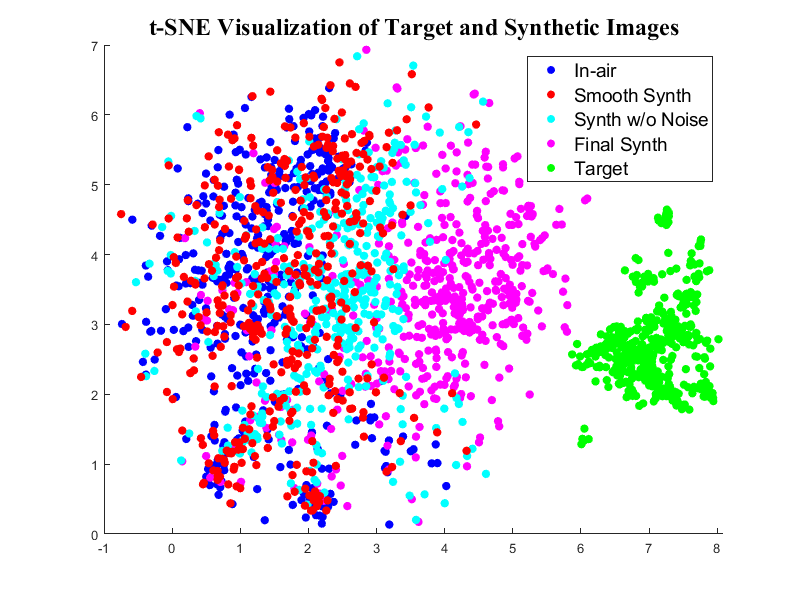}
    \label{fig:5_b}
}
\caption{\textbf{Quantitative Evaluation on Synthesis Components:} With the addition of the synthesis components, the FID scores show a decreasing trend.}
\label{fig:5}
\end{figure}

To evaluate the complete synthesis of target environments, we selected the underwater sequences: the \textbf{EASI-T} sequence, sourced from the EASI dataset \cite{yang2023knowledge}, which was created in an indoor water tank with baby powder used to adjust turbidity and the \textbf{Lake-C} sequence, a simulated cemetery scene captured in a lake, sourced from \cite{joshi2019experimental}, and the \textbf{AQUALOC} sequence from \cite{ferrera2019aqualoc}. Qualitative examples are shown in Fig. \ref{fig:4}. We visually demonstrate the difference between the smooth synthetic images without GAN-synthesized noise and forward scattering and the fully synthetic images. To assess the effectiveness of each component in the synthesis process, we conducted a quantitative evaluation using the GAN metric Frechet Inception Distance (FID), where a lower score indicates a closer resemblance to the target images. We compared the FID scores of four groups against real underwater images: in-air RGB images ($J_c$), smooth synthetic images without noise and forward scattering (Smooth), synthetic images without GAN noise ($I_c$), and fully synthesized images ($I^{Noise}_c$). Each group consisted of 500 randomly sampled images, with the results presented in Fig. \ref{fig:5}. We observe a decreasing trend in the FID scores as components are added, as illustrated in Fig. \ref{fig:5_a}. Using the EASI-T sequence as an example, we employed t-SNE to visualize the domain shift, which clearly shows a progression from left to right towards the target images in Fig. \ref{fig:5_b}. Notably, adding forward scattering and image noise components moves the synthetic images closer to the target domain. While a domain gap remains due to scene content differences between the in-air and underwater datasets, the t-SNE progression, particularly after incorporating image noise, clearly illustrates that modelling noise is both necessary and effective for image synthesis.

\begin{table}[tp]
    \centering
    \caption{Ablation study on the effect of forward scattering and noise}
    \begin{tabular}{|c|c|c|c|c|}
        \hline
        \multicolumn{3}{|c|}{\textbf{Synthetic Components}} & \multicolumn{2}{c|}{\textbf{Performance Metrics}} \\ \hline
      \textbf{G. Noise} & \textbf{GAN Noise} & \textbf{F. Scattering} & \textbf{M. Num} & \textbf{M. Rate} \\ \hline
      &  &  & 126.9 & 0.579\\ \hline
       &  & \checkmark & 104.4 & 0.468\\ \hline
      \checkmark &  &  &135.6 & 0.641 \\ \hline 
      \checkmark &  & \checkmark & 147.5  & 0.682\\ \hline
       & \checkmark&  & 152.3 & 0.642\\ \hline
      \checkmark & \checkmark & \checkmark & 163.1 & 0.765\\ \hline
       &  \checkmark& \checkmark & \textbf{191.3}&  \textbf{0.780} \\ \hline
    \end{tabular}
    \label{tab:2}
    \vspace{-4mm}
\end{table}

\subsection{Feature Extraction and Matching Results}
Using synthetic underwater images as the medium for knowledge distillation, the transferred model demonstrates enhanced performance in feature extraction and matching on images from the target environment. To showcase the generalization of our training framework, we employ not only the SuperPoint \cite{detone2018superpoint} as the teacher model, as in \cite{yang2023knowledge} but also the ALIKE model from \cite{zhao2022alike} as the teacher model. We use the models to extract $500$ feature points from $480 \times 640$ images.  Due to the lack of groundtruth in real underwater sequences, we apply RANSAC to estimate the transformation between image frames with 5-image intervals and filter out mismatches. We evaluate performance using \textbf{Matching Num} (M. N), defined as the number of correct matches \(N_{\text{match}}\) after RANSAC filtering, and \textbf{Matching Rate} (M. R), calculated as \(N_{\text{match}} / N_{\text{found}}\), where \(N_{\text{found}}\) denotes the total number of matches identified by the descriptors. The results are averaged over 100 randomly sampled image pairs from the sequences, as shown in Table~\ref{tabel1}. We use \textbf{ALIKE(KD)} and \textbf{SP(KD)} to represent the models trained by our proposed knowledge distillation framework on the corresponding sequences. Additionally, to demonstrate the effectiveness of the proposed model and its performance compared to UFEN, which was trained with randomly sampled water parameters and image white noise, we also performed knowledge distillation on SuperPoint with a binarization layer, labelled as \textbf{SP(KD-B)}, for a fair comparison with UFEN, whose descriptor is in binary format. The results demonstrate the effectiveness of the proposed knowledge distillation framework, which supports various in-air models and descriptor formats.

To evaluate the effectiveness of incorporating forward scattering and image noise in our synthetic image model, we conducted an ablation study using the knowledge-distilled SuperPoint model on the EASI-T sequence (Table \ref{tab:2}). Both Gaussian image noise (G. Noise) and GAN-synthesized noise (GAN Noise) enhance feature extraction and matching performance, with GAN Noise yielding superior results. Incorporating forward scattering (F. Scattering) further boosts performance. When both GAN-synthesized noise and forward scattering are considered, the model achieves the best performance. This study highlights the importance of accounting for forward scattering and image noise in underwater feature extraction and matching. 

\begin{table}[t]
\centering
\caption{Comparison of underwater image synthesis methods}
\begin{tabular}{|c|c|c|c|c|}
\hline
\multirow{2}{*}{\textbf{Synthesis Method}} & \multicolumn{2}{c|}{\textbf{Lake-C}} & \multicolumn{2}{c|}{\textbf{EASI-T}} \\ \cline{2-5} 
 & \textbf{M. Num} & \textbf{M. Rate} & \textbf{M. Num} & \textbf{M. Rate} \\ \hline
WaterGAN  & 251.9 & 0.816 & 140.8  & 0.649  \\ \hline
UWNR & 198.5 & 0.707 & 104.4 & 0.457 \\ \hline
\textbf{Ours} & \textbf{275.2} & \textbf{0.904}& \textbf{191.3}&  \textbf{0.780}\\\hline
\end{tabular}
\vspace{-4mm}
\label{tab:synth_comp}
\end{table}

Additionally, to demonstrate the effectiveness of our synthetic method, we compare our GAN-based synthesis with two existing approaches: WaterGAN~\cite{li2017watergan} and UWNR~\cite{ye2022underwater} in Tab.\ref{tab:synth_comp}. WaterGAN simulates backscattering generated from random input noise and introduces image noise during generation, but does not model forward scattering. While it brings performance gains for knowledge distillation, our method achieves better results. UWNR simulates the visual appearance of underwater scenes using natural light field rendering, producing visually realistic images. However, it does not account for either image noise or forward scattering, resulting in limited performance in feature transfer. These comparisons highlight the effectiveness of explicitly modeling both image noise and forward scattering in the synthesis pipeline.

\subsection{Downstream Application: Underwater VSLAM Results}
\begin{figure}[b]
\centerline{\includegraphics[width=0.45\textwidth]{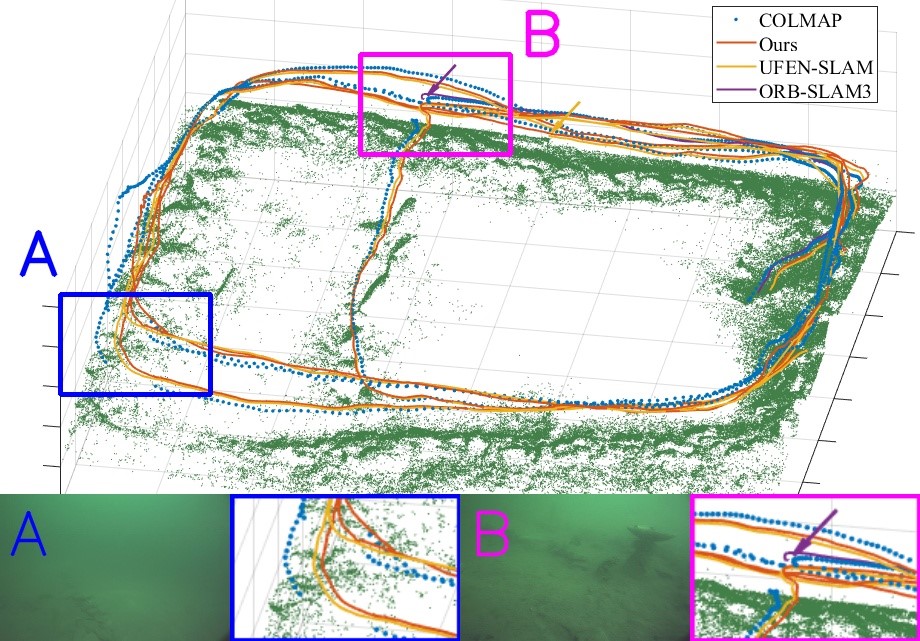}}
\caption{\textbf{Visualization of VSLAM Results:} Our method, with enhanced features, tracks the full trajectory. COLMAP reconstruction has gaps in blurry regions, while UFEN-SLAM and ORB-SLAM3 fail to track the full trajectory.}
\label{fig:8}
\end{figure}

We further utilized the transferred feature extraction networks, trained using our method, in downstream VSLAM tasks. The knowledge distillation of the SuperPoint pre-trained model was performed using our GAN-synthesized images as the medium, with a binarization layer attached at the end of the descriptor. The model \textbf{SP KD-B} mentioned above, transferred through the GAN-synthesized images, was integrated into the front end of the ORB-SLAM3 framework, hereafter referred to as \textbf{Ours}, with descriptors converted into a 256-bit binary format to comply with the ORB-SLAM3 standard. The VSLAM performance on the real underwater sequence Lake-C is presented in Fig. \ref{fig:8}. Due to the absence of groundtruth for highly turbid underwater sequences, we utilized the Structure-from-Motion method, COLMAP, to reconstruct the Lake-C sequence. However, there are reconstructed frame gaps due to poor matching in blurry regions, as illustrated by image samples from scenes A and B, highlighting substantial reconstruction errors. Both ORB-SLAM3 and UFEN-SLAM failed to track the entire trajectory due to high turbidity and flowing particles, whereas our method maintained tracking throughout the sequence. Similar results were observed on the EASI-T dataset, where even ORB-SLAM3 was unable to initialize.

\begin{table}[t]
\centering
\caption{ATE (m) comparison with different models}
\begin{tabular}{|c|c|c|c|c|c|}
\hline
\textbf{Model} & \textbf{01} & \textbf{02} & \textbf{03} & \textbf{05} & \textbf{06} \\
\hline
SP (w/o KD-B)     & \textemdash & \textemdash & \textemdash & 0.139 & 0.0260 \\
\hline
UFEN            & 0.143 & \textbf{0.153} & 0.0272 & 0.124 & 0.0217 \\
\hline
Ours            & \textbf{0.118} & 0.157 &\textbf{0.0259}  &\textbf{0.111}  & \textbf{0.0211} \\
\hline
\end{tabular}
\vspace{-4mm}
\label{tab:ate}
\end{table}

Meanwhile, we conducted VSLAM experiments on the AQUALOC dataset to evaluate the transferred model quantitatively. Sequences 04 and 07 were excluded due to consistent tracking failures across all methods, caused by severe lens occlusion and motion-induced blur.  To directly reflect the performance of front-end feature detection and matching, we disabled both loop-closure detection and motion mode tracking in the SLAM framework. Feature correspondences were established purely based on descriptor similarity using nearest neighbour matching. We compared the Absolute Trajectory Error (ATE) obtained from VSLAM using three models, a fine-tuned SuperPoint model with descriptor binarization but without knowledge distillation (SP w/o KD-B), UFEN, and our proposed method (SP KD-B). The results are summarized in Tab.~\ref{tab:ate}, indicating that the SuperPoint model without knowledge distillation exhibits unstable tracking performance and larger errors. In contrast, the transferred model trained with our proposed synthesis method, which incorporates both marine snow noise and forward scattering, demonstrates superior performance.

\section{Conclusion}
In this paper, to improve the feature matching performance of the in-air models in underwater environments, we propose a knowledge distillation framework that adapts in-air feature extraction models to underwater environments using GAN-synthesized images. However, a potential limitation of our method lies in extremely low-texture underwater environments, where the absence of visual cues may hinder matching robustness. Future work may explore integrating inertial or acoustic sensing to enhance performance under such conditions.
\bibliographystyle{IEEEtran}
\bibliography{IEEEabrv, ref}
\end{document}